\documentclass[letterpaper, 10 pt, conference]{ieeeconf}  % Comment this line out if you need a4paper

\IEEEoverridecommandlockouts                              % This command is only needed if 
\usepackage[colorlinks = true,
            bookmarks=true,
            linkcolor = magenta,
            urlcolor  = magenta,
            citecolor = cyan,
            anchorcolor = magenta]{hyperref}
\usepackage{url}            % simple URL typesetting
\usepackage{booktabs}       % professional-quality tables
\usepackage{amsfonts}       % blackboard math symbols
\usepackage{nicefrac}       % compact symbols for 1/2, etc.
\usepackage{microtype}      % microtypography
\usepackage{multicol,lipsum}
\usepackage{subcaption}
\usepackage[skip=2pt,font=footnotesize]{caption}
\usepackage{comment}
\usepackage{booktabs, tabularx} % for professional tables
\usepackage{makecell}
\usepackage{multirow}
\usepackage{wrapfig}
\usepackage{times}
\usepackage{url}
\usepackage{algorithm}
\usepackage{algorithmic}
\usepackage{graphicx}
\usepackage{mathtools}
\usepackage{amsmath}
\usepackage{amssymb}
\usepackage{pifont}% http://ctan.org/pkg/pifont
\usepackage{array}
\usepackage{soul}

\newcolumntype{P}[1]{>{\arraybackslash}p{#1}}
\newcolumntype{M}[1]{>{\arraybackslash}m{#1}}

\title{\LARGE \bf
ReorientDiff: Diffusion Model based Reorientation for Object Manipulation
}

\author{Utkarsh A. Mishra and Yongxin Chen% <-this % stops a space
\thanks{Utkarsh A. Mishra and Yongxin Chen are affiliated to the Institute for Robotics and Intelligent Machines~(IRIM), Georgia Institute of Technology
        {\tt\small umishra31@gatech.edu}, {\tt\small yongchen@gatech.edu}}%
}

\begin{document}

\maketitle
\thispagestyle{empty}
\pagestyle{empty}

%%%%%%%%%%%%%%%%%%%%%%%%%%%%%%%%%%%%%%%%%%%%%%%%%%%%%%%%%%%%%%%%%%%%%%%%%%%%%%%%
\begin{abstract}
% The ability to manipulate objects in a desired configurations is a fundamental requirement for robots to complete various practical applications. While certain goals can be achieved by picking and placing the objects of interest directly, object reorientation is needed for precise placement in most of the tasks. In such scenarios, the object must be reoriented and re-positioned into intermediate poses that facilitate accurate placement at the target pose. To solve this, we propose ReorientDiff, a diffusion model based reorientation planner which uses visual inputs from the scene and goal-specific language prompts to plan intermediate reorientation poses. The scene and the language-task is mapped to a joint scene-task representation feature space which is further used to condition a diffusion model. The diffusion model samples based on the representation using classifier-free guidance and then uses gradients of learned feasibility-score models for implicit iterative pose-refinement. After obtaining the sampled poses, a motion planner is used to execute the two-step operation. We achieve 95.1\% success rate and show our method's performance in simulation for a set of YCB-objects and a suction gripper.

The ability to manipulate objects in desired configurations is a fundamental requirement for robots to complete various practical applications. While certain goals can be achieved by picking and placing the objects of interest directly, object reorientation is needed for precise placement in most of the tasks. In such scenarios, the object must be reoriented and re-positioned into intermediate poses that facilitate accurate placement at the target pose. To this end, we propose a reorientation planning method, ReorientDiff, that utilizes a diffusion model-based approach. The proposed method employs both visual inputs from the scene, and goal-specific language prompts to plan intermediate reorientation poses. Specifically, the scene and language-task information are mapped into a joint scene-task representation feature space, which is subsequently leveraged to condition the diffusion model. The diffusion model samples intermediate poses based on the representation using classifier-free guidance and then uses gradients of learned feasibility-score models for implicit iterative pose-refinement. The proposed method is evaluated using a set of YCB-objects and a suction gripper, demonstrating a success rate of 95.2\% in simulation. Overall, we present a promising approach to address the reorientation challenge in manipulation by learning a conditional distribution, which is an effective way to move towards generalizable object manipulation. More results can be found on our website: \url{https://utkarshmishra04.github.io/ReorientDiff}.
\end{abstract}

%%%%%%%%%%%%%%%%%%%%%%%%%%%%%%%%%%%%%%%%%%%%%%%%%%%%%%%%%%%%%%%%%%%%%%%%%%%%%%%%
\section{Introduction}
\label{sec:introduction}

\begin{figure*}[t]
\centering
\includegraphics[width=0.9\linewidth]{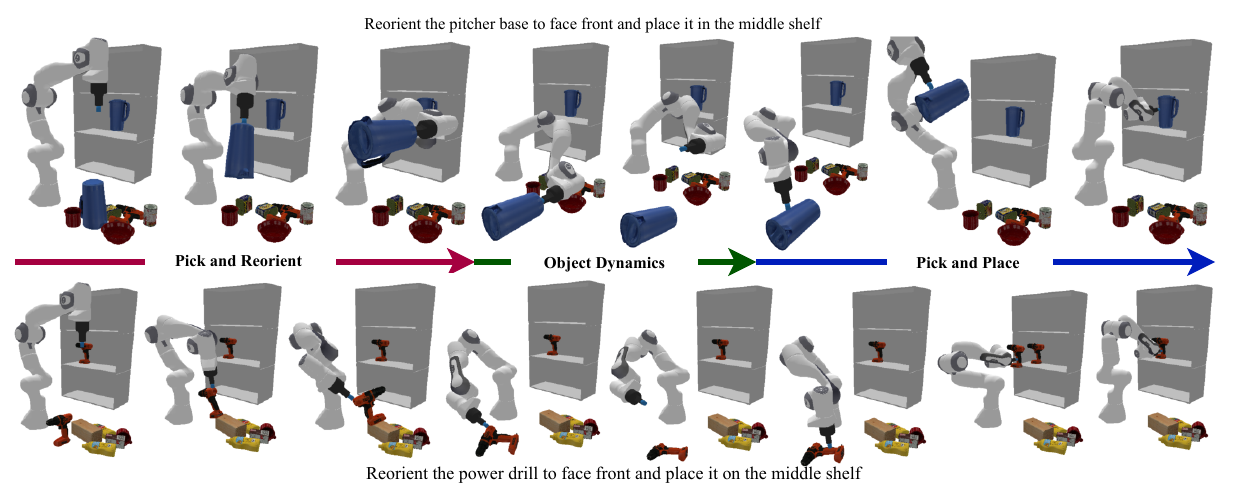}
    \caption{\textbf{Reorientation for precise target placement} The above figure represents the phenomenon of reorientation in which an object from a cluttered file has to be placed precisely in a shelf~(target position shown). As the object cannot be directly placed at the target location, our proposed method, ReorientDiff, samples a reorientation pose using a learned conditional distribution by a diffusion model. Such a proposed reorientation pose acts as a transition for facilitating successful placement. We also consider and take advantage of the object dynamics, as introduced by Wada~\textit{et al.}~\cite{wada2022reorientbot}, by which we ensure that un-grasping an object in an unstable pose will eventually allow the object to settle at some favourable pose.}
    \label{fig:train_pitcher_drill}
    \vspace{-1em}
\end{figure*}

Rearranging objects into specific poses is a fundamental task. It's not only essential for everyday activities at home but also plays a critical role in industrial applications like packing and assembly lines. Performing such a task requires extracting object information from visual-sensor data and planning a pick-place sequence~\cite{zeng2021transporter, tang2022selective}. While a single-step pick-place sequence is a viable solution, placing the object at a specific position and orientation is not always feasible. \emph{Reorientation} is an effective strategy when successfully changing an object's pose allows its placement at the target pose~\cite{wada2022reorientbot}. Such a strategy ensures feasible intermediate transition poses in scenarios without common grasps between the current pose and an object's desired placement pose.

The problem of finding reorientation poses is traditionally approached via rejection sampling based on finding successful grasps between the current pose-intermediate pose and intermediate pose-target pose. While previous classical approaches achieve this by using trajectory planners~\cite{kuffner2000rrt} to plan motion from the current pose to the desired pose via diverse candidate intermediate poses, such an exhaustive search is expensive on time and is limited by choice of the number of intermediate pose options. Recently, there have been efforts to improve the reorientation process via a data-driven rejection sampling solution using learned models~\cite{wada2022reorientbot} that predict the feasibility score of an intermediate pose w.r.t. feasible grasps in the current and target pose. 
% and help in early evaluation for rejection sampling. 
While their method improves the success rate and planning time, the algorithm requires processing significantly large number of candidate random samples and specifying the target object's placement pose. The former limits \emph{scalability}, and the latter challenges \emph{generalizability}. Lately, with the advances in language descriptor foundation models like CLIP~\cite{radford2021learning}, which projects images and texts to a common feature space, target object specifications can be directly correlated between visual information and suitable language commands, thus empowering human-robot interaction. This motivated us to explore grounding the problem statement of reorientation on language and hence embed semantic knowledge of the task with the spatial structure of the scene~\cite{shridhar2022cliport}.

In this paper, we introduce ReorientDiff, a diffusion model based \emph{generative} method to restructure the reorientation pose generation pipeline as a conditional distribution learning problem. Such a method enables us to directly sample feasible reorientation poses without rejection sampling, thus improving \emph{scalability}. Our contributions can be summarized as follows:

\textbf{Learning a distribution of intermediate poses:} For a given pile of objects, a target object, and its target placement location, we formulate a conditional distribution of feasible intermediate poses. As compared to rejection sampling using random prior, our approach aims at providing a learned prior to efficiently sample high-quality reorientation poses. Leveraging the multi-modality of diffusion models, this distribution encompasses all poses reachable from both the current pose and the target pose. 
% Notably, this formulation implicitly accounts for the effects of gravity, as illustrated in~\autoref{fig:train_pitcher_drill}.

\textbf{Flexibly sampling based on possible grasp poses:} It is necessary to make sure that the grasp poses w.r.t. object is constant during one pick-place transition. To achieve this, we flexibly sample intermediate poses from the learned distribution based on feasible grasp poses using classifier guidance via pre-trained success classifiers~\cite{mousavian20196dof, wada2022reorientbot}. Such models implicitly refine sampled pose and operate individually for both transitions during reorientation. Hence, the learned distribution can be used for any possible grasp pose based on kino-dynamic feasibility directly at inference.

\textbf{Representing target placement location via natural language:} We leverage CLIP~\cite{radford2021learning} to generate information embeddings from visual input and task descriptions in natural language.  We further use these embeddings as conditions for learning the conditional distribution. While this has been explored in recent literature~\cite{shridhar2022cliport}, we see this as a substantial improvement over the baseline.

In the proposed approach, we combine a generic classifier-free conditional sampling~\cite{ho2022classifier} with classifier-guided sampling~\cite{dhariwal2021diffusion} to sample from diffusion models. To validate the performance of ReorientDiff, we consider reorientation of objects in the YCB dataset~\cite{calli2015ycb} that are feasible for suction grippers. For each selected object, we choose suitable locations on multiple shelf levels and target orientations.

\section{Related Work}
\label{sec:relatedwork}

\textbf{Object manipulation and reorientation.} Finding the grasp pose that is feasible for both the current and target location is a widely employed strategy for pick-and-place operations~\cite{mahler2017dex, mahler2018dex, mahler2019learning}. Such problems are usually solved in two steps: deciding an appropriate placement pose (within a region of interest) and searching for common grasps. In order to ensure feasible target placement, prior works have mostly relied on known object geometries~\cite{mahler2017dex, mahler2018dex}, vision-based object representation~\cite{zeng2021transporter, seita2021learning} or using segmentation and depth maps of the pre-specified target object~\cite{pinto2016supersizing, zeng2022robotic, wada2022reorientbot}. These strategies have led to several object rearrangement methods~\cite{shridhar2022cliport, shridhar2022perceiver, tang2022selective}. Unlike most prior works that consider the availability of common grasps by default, for complex manipulation scenarios where there are no common grasps, \emph{reorientation} becomes mandatory. The object needs to be reoriented to an intermediate pose and \emph{regrasped} to place it at the target location. Such a scenario has been traditionally tackled via rejection sampling strategies and recently improved via regression-based methods. We also aim to develop a learning-based method.  

\textbf{Learning for object manipulation.} While prior works have pre-dominantly incorporated trajectory planners~\cite{kuffner2000rrt}, they have employed learning strategies to decide the target object and its placement pose as discussed in the previous subsection. Additionally, task descriptions as natural language have been very effective for generalized pick-place tasks in planar tabletop~\cite{shridhar2022cliport} and 3D~\cite{shridhar2022perceiver} manipulation. Such language descriptions can be embedded into the learning pipeline via foundation models like CLIP~\cite{radford2021learning}, which encodes visual and language information into a common representation space. This has been further extended towards language-conditioned object rearrangement planning~\cite{liu2022structdiffusion, liu2022structformer} and supplying high-level instructions for long-horizon planning~\cite{ahn2022can}.

Recently, \emph{reorientation} problems have been solved by planning to reorient objects using extrinsic supports~\cite{cheng2021learning, xu2022planar}, which enables them to re-grasp the object in a desired way. The above methods are regression-based and limited to modeling only one solution pose. Such approaches cannot cater to the multiple possible solutions of the same problem. In such a case, rejection sampling is still beneficial and can be performed using learned feasibility prediction models~\cite{wada2022reorientbot}. We want to develop a pipeline that can still learn about all feasible poses without analyzing extensive random samples.

\textbf{Generative models for object manipulation.} For pick-and-place and reorientation tasks, there can be multiple feasible grasps and reorientation poses respectively. Hence, generative models offer an option to learn them as conditional distributions. Prior works have explored VAE for planning grasps~\cite{mousavian20196dof} using visible point-cloud of objects. In this direction, diffusion models have been shown to be advantageous for robotics~\cite{janner2022diffuser, ajay2022conditional, chi2023diffusion, mishra2023generative, xian2023unifying}. Recent works have demonstrated the multi-modal distribution learning using diffusion models for finding target poses~\cite{liu2022structdiffusion, simeonov2023shelving} and learning policies~\cite{janner2022diffuser, ajay2022conditional, chi2023diffusion}. In addition to such properties, we also plan to leverage the flexible sampling and conditioning strategies offered by diffusion models to incorporate additional conditions at inference without re-training.

\begin{figure*}[t]
\centering
\includegraphics[clip=True, trim={0 0.5cm 0 0}, width=0.9\linewidth]{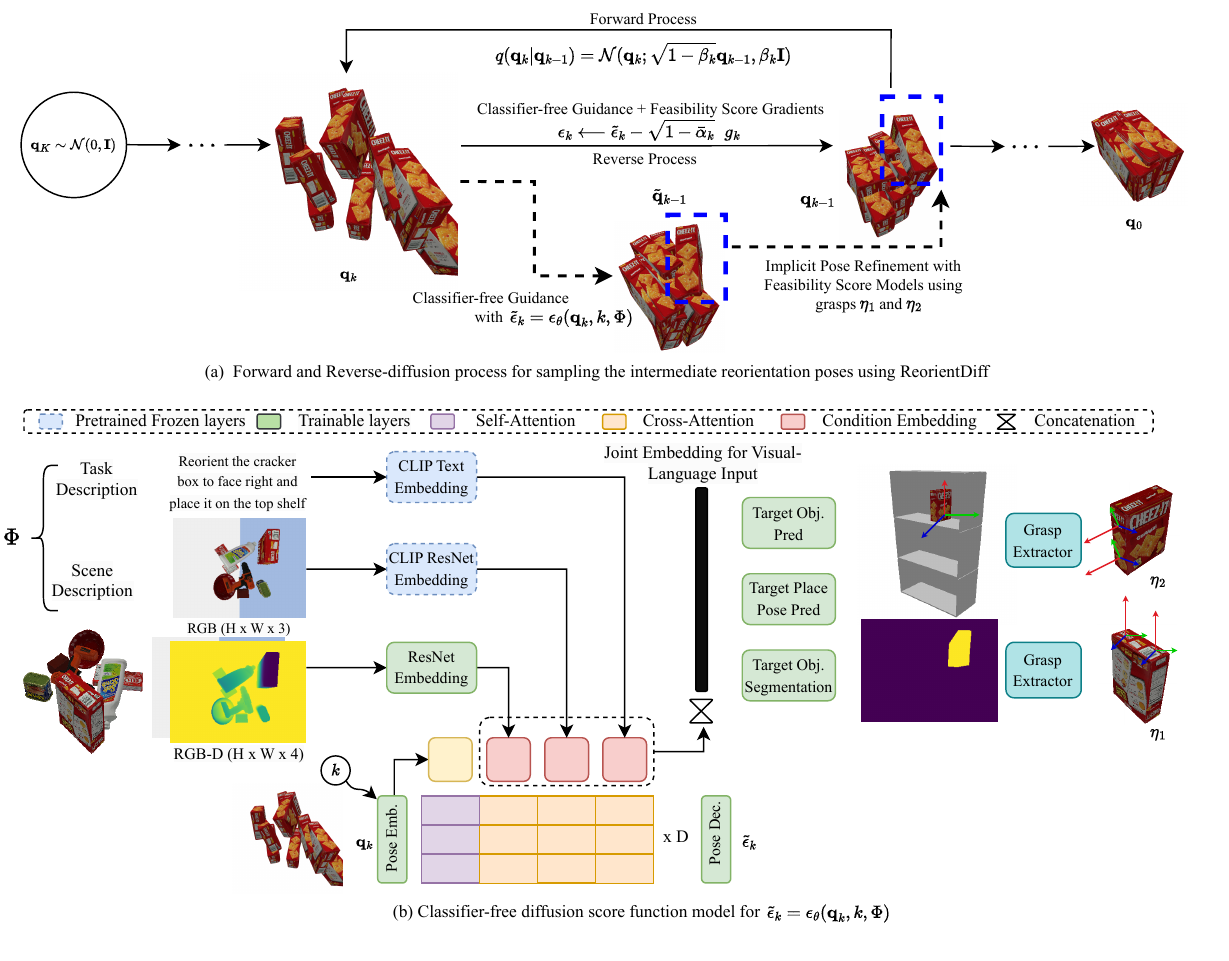}
    \caption{\textbf{Method Overview} (a) \textbf{Forward and reverse diffusion process.} ReorientDiff uses a combination of classifier-free guidance with classifier-based implicit refinement to sample from the learned distribution of intermediate poses. It ensures high-success feasibility with minimal variance by guiding the scene-task conditioned sampling using feasibility score gradients. (b) \textbf{Conditioned score function.} ReorientDiff learns the target distribution of feasible reorientation poses conditioned on the scene (pile of objects) and task (language prompt) jointly represented as $\Phi$. We use the pre-trained frozen CLIP text and image embeddings to formulate a joint embedding, trained end-to-end to encode information about placement pose, target object and current pose. Further, the current pose and target poses are processed to obtain feasible grasps~($\eta_1$ and $\eta_2$), which are used to calculate the feasibility gradients~$g_k$ in (a). The joint embedding is used as a sequence to condition the transformer-based score network $\epsilon_\theta(\textbf{q}_k, k, \Phi)$ via cross-attention to obtain the classifier-free score estimate in (a).}
    \label{fig:embedding}
    \vspace{-1em}
\end{figure*}

\section{Preliminary: Diffusion Models}
\label{sec:preliminary}

Consider samples $x_0$ from an unknown data distribution $q(x_0)$; diffusion models~\cite{ho2020denoising} learn to estimate the distribution by a parameterized model $p_\theta(x_0)$ using the given samples. The procedure is completed in two steps: the forward and the reverse diffusion processes. The former continuously injects Gaussian noise in $x_0$ to create a Markov chain with latents $x_{1:K}$ following transitions:
\begin{equation}
    q(x_{1:K}|x_0) = \prod_{k=1}^K q(x_k|x_{k-1}),
\end{equation}
where $q(x_k|x_{k-1})~=~\mathcal{N}(x_{k};\sqrt{1-\beta_k}x_{k-1}, \beta_k \mathbf{I})$ is the per-step noise injection following variance schedule $\beta_1, \dots, \beta_K$. This leads to the distribution $q(x_k|x_0)~=~\mathcal{N}(x_{k};\sqrt{\bar{\alpha}_k}x_{0},~(1-\bar{\alpha}_k)~\mathbf{I})$ following notations introduced in~\cite{song2020denoising} as $\alpha_k = 1 - \beta_k$ and $\Bar{\alpha}_k = \prod_{i=1}^k \alpha_i$. Note that $\bar \alpha_K \approx 0$ and thus $x_K \sim \mathcal{N}(0, \mathbf{I})$. The reverse diffusion learns to denoise the data starting from $x_K$ and following $p_\theta(x_{k-1}|x_k)~=~\mathcal{N}(x_{k-1};\mu_\theta(x_{k}, k), \beta_k \mathbf{I})$ where
\begin{equation}\label{eq:mu}
    \mu_\theta(x_{k}, k) = \frac{1}{\sqrt{\alpha_k}} \Big( x_k - \frac{\beta_k}{\sqrt{1 - \bar{\alpha}_k}} \epsilon_\theta(x_k, k) \Big).
\end{equation}
The parameterized model $\epsilon_\theta(x_k, k)$ is called the score-function, and it is trained to predict the perturbations and the noising schedule by the score-matching objective~\cite{song2020score}
\begin{equation}
    \arg\min_\theta \mathbb{E}_{x_0\sim q, \epsilon\sim \mathcal{N}(0, \mathbf{I})} \Big[\| \epsilon - \epsilon_\theta( \sqrt{\bar{\alpha}_k}x_{0} + \sqrt{1-\bar{\alpha}_k} \epsilon,k) \|^2\Big]
\end{equation}
In particular, such a score function represents the gradients of the learned probability distribution as
\begin{equation}
    \nabla_{x_k} \log p_\theta(x_k) = - \frac{1}{\sqrt{1-\bar{\alpha}_k}} \epsilon_\theta(x_k, k).
\end{equation}

% \begin{figure*}[t]
% \centering
% \includegraphics[width=\linewidth]{figures/forward_rev_process.pdf}
%     \caption{\textbf{Forward and Reverse Diffusion Process} The above figure shows the forward diffusion and the reverse denoising and sampling process of ReorientDiff. As described in~\autoref{sec:reorientdiff}, following classifier-free guidance will result in high-likelihood samples with high-variance in terms of success feasibility of the samples. Using the feasibility score gradients, we realize an implicit iterative pose refinement, as marked by the blue box in the figure. This significantly decrease variance and ensure high success feasibility of the samples.}
%     \label{fig:reorientdiff}
%     \vspace{-1em}
% \end{figure*}

\section{Reorientation}
\label{sec:reorientation}
% \subsection{Reorientation}
% \chen{this section introduces the overall pipeline. we can briefly mention how generative model is used in this pipeline}

Reorientation consists of solving two problems simultaneously, finding a pose that is reachable from the current pose and, after the effect of gravity, results in a pose that makes placement at the target pose achievable~(as shown in \autoref{fig:train_pitcher_drill}). 
% \chen{what? This makes} the problem challenging. 
Once we have an estimate of the current and target pose, it is intuitive that there will be a set of poses that will satisfy reorientability. However, only a small subset of such reorientable poses will be valid with provided kino-dynamic constraints on grasp poses. Identifying a candidate sample from this subset by either brute force sampling or optimization is computationally expensive and has to be done for every new scenario. 

To circumvent the above challenges, we propose a generative modeling approach to sample from the subset of valid reorientation poses. More specifically, our method learns the distribution of all reorientable poses using a conditional diffusion model and use classifiers to guide sampling towards valid poses directly during inference based on provided grasp poses. Hence, we divide the problem into three segments: i) regression-based end-to-end learning for finding the target object and placement pose from the scene and task description~(scenario), ii) learning the distribution of all reorientable poses for a given scenario once the object specifications are known and iii) learning grasp feasibility classifiers for selecting only the valid reorientation poses. To achieve this, we discuss our formulation for constructing scene-task representation, calculating grasp poses from object poses and learning grasp feasibility classifiers below. The diffusion model training and inference is discussed in the next section.

\subsection{Constructing Generic Scene-Task Representations}

A scene-task representation is a compact embedding of all available information present in the scene and specified by the user. We define a scene as the location and occupancy of the place from where a target object should be picked and a task as the language prompt containing the descriptions for selecting the target object and deciding placement poses. A top-down RGB-D camera provides an image~$\mathcal{I}\in\mathbb{R}^{H\times W\times3}$ and a heightmap~$\mathcal{H}\in\mathbb{R}^{H\times W \times 1}$ as the description of the pile. For learning the semantic and spatial embeddings~\cite{shridhar2022cliport, shridhar2022perceiver}, we use pre-trained CLIP foundation model and obtain semantic embeddings from the image~$\mathcal{I}$ and language~$\mathcal{L}$. We sequence the embeddings with the 
% trained ResNet50~(spatial) 
spatial embeddings for target object segmentation 
% with RGBD~($[\mathcal{I}, \mathcal{H}]$) 
to get a joint embedding sequence~$\Phi$ as generic scene-task representation as shown in~\autoref{fig:embedding}(b). 
The embedding is further used to predict the target object and the final placement pose.

\subsection{Sampling Grasp Poses}

We generate grasp poses by following the classical approach of converting the heightmap into a point cloud representation and eventually to a point-normal representation~\cite{wada2022reorientbot}. The predicted target object segmentation of the scene is then used to obtain the surface normals of the target object. After performing an edge masking using the Laplacian of the surface normals, the remaining point-normals on the surface are feasible grasp poses. While we sample grasp poses~$\eta_1$ for picking the object from the pile in the aforementioned manner, we assume that we have the mesh of the selected object
% \footnote{\chen{this could be too much for a footnote. why not include it in the main paper?} \addressed{no space} For complicated manipulation, as one shown in this work, the whole shape of the object plays an important role. While previous single-step pick and place tasks have all the necessary information about the object from a singe-view perspective, such an information is incomplete for reorientation tasks~(multi-step) which require knowledge of parts of the object not visible in the current pose. Previous related works~\cite{wada2022reorientbot, cheng2021learning, xu2022planar, wermelinger2021grasping} have proposed their approach under similar assumptions.} 
for sampling grasp poses~$\eta_2$ for placing the object at the predicted pose.

\subsection{Feasibility Score Models}
% \chen{is feasibility score model a good name? we already have score function}
Following prior works~\cite{mousavian20196dof, wada2022reorientbot, liu2022structformer}, a feasibility prediction model is important for early-evaluation and rejection of unfavorable samples. Such a feasibility model predicts the probability of success of a given grasp pose in successfully grasping an object in some candidate pose for a specified scene representation. The phenomenon of grasp success evaluation in dynamic reorientation pose, as addressed by~\cite{wada2022reorientbot}, is particularly interesting for our setup. Modelling dynamics for every object is indeed non-trivial and adds to the complexity; hence the feasibility model implicitly takes care of the dynamics of the object after deactivating the grasp. For checking feasibility or the probability of success~($y$) of sampled grasps for candidate reorientation poses~$\mathbf{q}$, we train two models: 
% \addressed{clarify what $y$ is}
\begin{itemize}
    \item For predicting success of reorientation from the current pose in a pile to a candidate pose given pick grasp poses~($\eta_1$) and scene representation, denoted as~$\mathcal{M}_1(y|\eta_1, \mathbf{q}, \Phi)$
    \item For predicting success of post-grasp deactivation pose from the candidate pose and placement grasp poses~($\eta_2$), denoted as~$\mathcal{M}_2(y|\eta_2, \mathbf{q}, \Phi)$
\end{itemize}

\section{ReorientDiff: Diffusion for Reorientation}
\label{sec:reorientdiff}
% \addressed{maybe we should include a diagram for reorientdiff}
We aim to generate intermediate reorientation poses for the target object, which enables successive placement at the desired pose and is reachable from the current pose.
% While prior works have used rejection sampling~\cite{wada2022reorientbot} and end-to-end supervised learning~\cite{xu2022planar, cheng2021learning}, 
We introduce a diffusion model-based approach to sample the most probable successful reorientation poses~($\mathbf{q}$) conditioned on the scene representation priors~($\Phi$), denoted as $p(\mathbf{q}|\Phi)$, which already contains the spatial and semantic information about the scene and the task. The denoising process can be further flexibly conditioned by sampling from modified distributions of the form
\begin{equation}
    \label{eq:dist}
    p_{h}(\mathbf{q}) \propto p(\mathbf{q}|\Phi) h(\mathbf{q}, \Phi),
\end{equation}
where $h(\mathbf{q}, \Phi)$ can represent several grasp success probability heuristics. By separating the grasp success from reorientation candidate sampling, the diffusion model trained for reorientation poses can be reused for varied selection of picking~($\eta_1$) and placement grasp poses~($\eta_2$).

\subsection{Classifier-free Conditional Pose Generation}

Following the distribution defined in ~\eqref{eq:dist}, we use classifier-free guidance~\cite{ho2022classifier} to sample high-likelihood reorientation poses for a particular scene-task representation. We train a score-network~\cite{song2020score}, $\epsilon_\theta(\mathbf{q}_k, \Phi) \propto \nabla_{\mathbf{q}_k} \log p(\mathbf{q}_k|\Phi)$ 
% \addressed{or $\nabla_{\mathbf{q}_k} \log p(\mathbf{q}_k| \Phi)$?}
, to denoise from~$\mathbf{q}_K \sim \mathcal{N}(\bf{0}, \mathbf{I})$ to possible reorientation poses~$\mathbf{q}_0$ from a $K$-step reverse diffusion denoising process. For each step, we calculate $\Tilde{\epsilon}_k$ as
\begin{equation}
    \Tilde{\epsilon}_k = \epsilon_\theta(\mathbf{q}_k, \Phi) + w_c \Big(\epsilon_\theta(\mathbf{q}_k, \Phi) - \epsilon_\theta(\mathbf{q}_k, \o)\Big)
\end{equation}
The scalar $w_c$ implicitly guides the reverse-diffusion towards poses that best satisfy the scene-task representations. Further, we calculate the successive samples for the next $(k-1)^{th}$ step using the DDIM~\cite{song2020denoising} sampling strategy and $\Tilde{\epsilon}_k$ as follows:
\begin{align}
\label{eq:update_diffusion}
    \tilde{\mathbf{q}}_{k-1} &\longleftarrow \sqrt{\Bar{\alpha}_{k-1}}\Big(\frac{\mathbf{q}_k - \sqrt{1 - \Bar{\alpha}_k} \;\;\Tilde{\epsilon}_k }{\sqrt{\Bar{\alpha}_k}}\Big) + \sqrt{1 - \Bar{\alpha}_{k-1}}\;\;\Tilde{\epsilon}_k
\end{align}
where, $\Bar{\alpha}_{k}$ is as described in~\autoref{sec:preliminary}.

\subsection{Feasibility Guided Pose Refinement}

We use the two feasibility-score prediction models ($\mathcal{M}_1$ and $\mathcal{M}_2$), which are pre-trained for predicting grasp feasibility for picking grasp, reorientation pose pairs and placement grasp, reorientation pose pairs, respectively. In such a case, the scores can be converted into probability distributions for each heuristic, defined as, for each $i=1,2$,
% \addressed{minus sign $-$ missing?}
\[
    h_i \equiv \ p(y=1|\eta_i, \mathbf{q}, \Phi)|_{\mathcal{M}_i} =  \text{exp}\Big(-(1 - \mathcal{M}_i(y|\eta_i, \mathbf{q}, \Phi))^2\Big) 
    % \\
    % h_2 \equiv \ p(y=1|\eta_2, q, \Phi)|_{\mathcal{M}_2} = \ \text{exp}\Big(-(1 - \mathcal{M}_2(y|\eta_2, q, \Phi))^2\Big)
\]

Following classifier-based guidance~\cite{dhariwal2021diffusion} formulation for the heuristics, the reverse diffusion can be formulated as: 
% \chen{I thought $h_1, h_2$ only depend on $\mathbf{q}_0$} 
\begin{multline}
\label{eq:new_dist}
    p_{h}(\mathbf{q}_k|\mathbf{q}_{k+1}, y, \Phi) \propto \\  p(\mathbf{q}_k|\mathbf{q}_{k+1},\Phi) \;\; p(y|\eta_1, \hat{\mathbf{q}}_0^k, \Phi)|_{\mathcal{M}_1} \ p(y|\eta_2, \hat{\mathbf{q}}_0^k, \Phi)|_{\mathcal{M}_2}
\end{multline}
where, $\hat{\mathbf{q}}_0^k$ is the sample proposed at diffusion step $k$ and defined as: 
% \chen{the noise below maybe incorporated by using a stochastic version of DDIM directly}
\begin{equation}
    \hat{\mathbf{q}}_0^k = \frac{\mathbf{q}_k - \sqrt{1 - \Bar{\alpha}_k} \;\;\Tilde{\epsilon}_k }{\sqrt{\Bar{\alpha}_k}}
\end{equation}
Considering Taylor first order approximations for heuristics and standard reverse process Gaussian $(\mu_\theta(\mathbf{q}_k, k, \Phi), \beta_k \mathbf{I})$ as described in~\autoref{sec:preliminary}, we get the new mean~($\mu_{\theta, h}(\mathbf{q}_k, k, \Phi)$) for the distribution~$p_{h}(\mathbf{q}_k|\mathbf{q}_{k+1}, y, \Phi)$ in \eqref{eq:new_dist} as: 
% \addressed{what's $\sigma$?} \addressed{missing sign?}
\begin{align}
     & \mu_{\theta, h}(\mathbf{q}_k, k, \Phi) \nonumber \\
     &= \mu_\theta(\mathbf{q}_k, k, \Phi) + \beta_k \sum_{i=1}^2 w_i\nabla_{\mathbf{q}_k} \log p(y|\eta_i, \mathbf{q}_k, \Phi)|_{\mathcal{M}_i} \nonumber \\
     &= \mu_\theta(\mathbf{q}_k, k, \Phi) - \beta_k \sum_{i=1}^2 w_i\nabla_{\mathbf{q}_k} \Big[1 - \mathcal{M}_i(y|\eta_i, \hat{\mathbf{q}}_0^k, \Phi)\Big]^2. \nonumber
\end{align}
In view of \eqref{eq:mu}, we then obtain the modified score
    \[
    \epsilon_k \longleftarrow \Tilde{\epsilon}_k - \sqrt{1 - \Bar{\alpha}_k} \ g_k
    \]
where $g_k = -\beta_k\sum_{i=1}^2 w_i\nabla_{\mathbf{q}_k} \Big[1 - \mathcal{M}_i(y|\eta_i, \hat{\mathbf{q}}_0^k, \Phi)\Big]^2$. We notice that injecting noise to $g_k$, as in stochastic DDIM, can slightly improve the performance. 
% Now, to incorporate the feasibility-score gradients into the standard score function, we calculate the gradients as follows:
% \[
%     g_k = -\beta_k\sum_{i=1}^2 w_i\nabla_{\mathbf{q}_k} \Big[1 - \mathcal{M}_i(y|\eta_i, \hat{\mathbf{q}}_0^k, \Phi)\Big]^2 + \sqrt{\zeta \sigma} \Tilde{q},
% \]
% and adapt the score function according to the following:
% \begin{align}
%     \epsilon_k &\longleftarrow \Tilde{\epsilon}_k - \sqrt{1 - \Bar{\alpha}_k} \;\; g_k
% \end{align}
We calculate the final $\mathbf{q}_{k-1}$ using the refined $\epsilon_k$ in \eqref{eq:update_diffusion}. A visual clarification of the forward and reverse diffusion is shown in~\autoref{fig:embedding}(a).

\begin{figure}[t]
\centering
\includegraphics[clip=True, trim={0 0.3cm 0 0}, width=\linewidth]{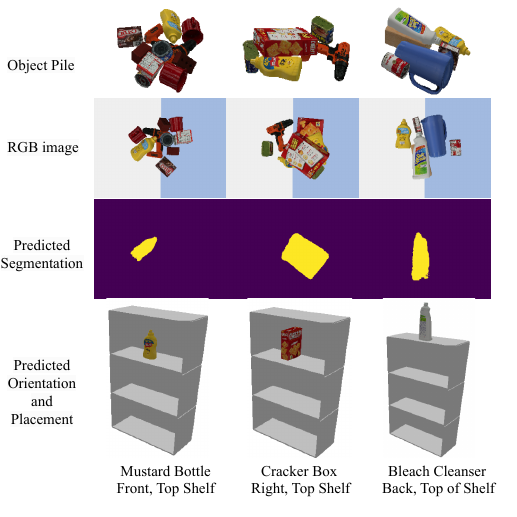}
    \caption{\textbf{Visual Analysis of Scene-Task Network Performance} The scene-task network maps the visual~(row $2$) image of the pile~(row $1$) and language~(bottom row) inputs to a feature space which is used to predict the placement location~(row $4$) and target object segmentation~(row $3$).}
    \label{fig:networkeval}
    \vspace{-1em}
\end{figure}

\begin{figure*}[t]
\includegraphics[clip=true, trim={0 0.3cm 0 0}, width=\linewidth]{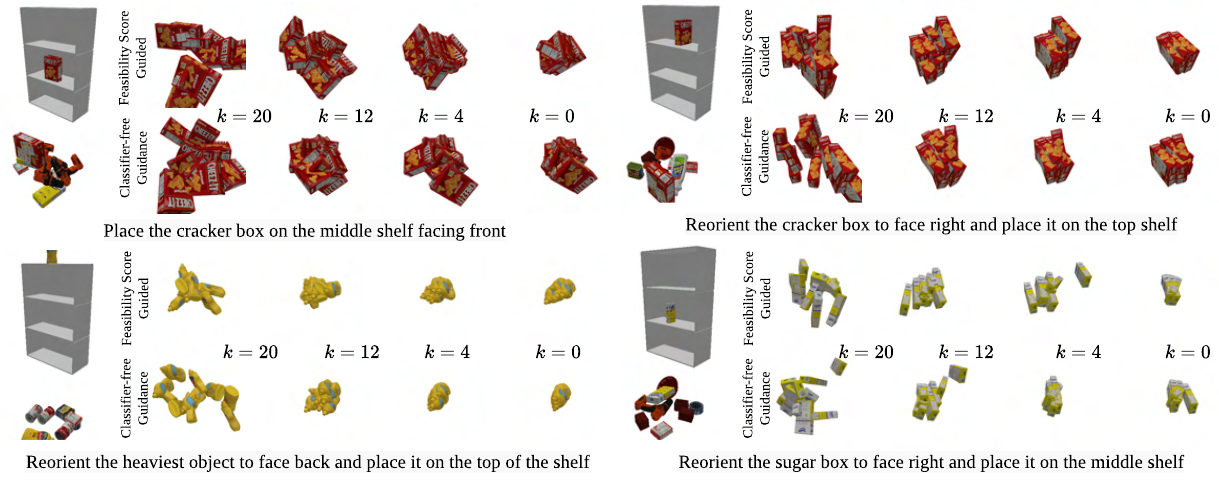}
    \caption{\textbf{Reverse Diffusion for Reorientation Pose Generation} The reverse sampling process for $4$ $k$-steps at $k=20, 12, 4, 0$ for $K=20$ 
    % \addressed{$k=100$ to $k=0$?}
    in four different scene-task scenarios comprising of the Cracker Box, Mustard Bottle and Sugar Box in different target orientations are shown above. The scenes are shown in the left-side of every sub-figure and consists of the pile with the target object and the predicted placement location on the shelf. The language prompt defining each of the tasks is mentioned below each sub-figure. It consists of either the absolute (the object's name) or the relative~(heaviest/lightest) reference to the object and details about the target placement.}
    \label{fig:diffusionObjects}
\vspace{-1em}
\end{figure*}

\section{Results: Simulation}
\label{sec:resultssim}

Based on the environment setup as discussed in~\autoref{sec:reorientation}, we create datasets, train diffusion and feasibility score models and evaluate them in simulation.
% for proper placement conditions.

\subsection{Dataset Generation and Training}

We use PyBullet~\cite{coumans2021} and an OMPL~\cite{sucan2012ompl} based motion planner to solve for collision-free path between current pose and a candidate reorientation pose and from the reorientation pose to the ground-truth placement pose for diverse set of YCB-objects and target locations. We sampled approximately $40000$ candidate poses following Wada~\textit{et al.}~\cite{wada2022reorientbot}. The goal properties were converted into modular language instructions, and the success of pick and place for both the steps was recorded. The scene and task properties were used to construct the joint visual-language embedding space, which was further used to train the feasibility score models using binary success labels. Eventually, we train a conditional diffusion model using only the successful reorientation poses. Such a diffusion model is reusable for diverse set of grasp poses when combined with the feasibility score models.

\subsection{Performance Evaluation: Scene-Task Representation}

To evaluate the quality of the scene-task embedding network, we analyze the accuracy of the object selection and placement pose prediction along with the error in the predicted segmentation. We show a visual analysis in~\autoref{fig:networkeval} where the output segmentation and the predicted placement pose in the shelf are shown for three scenes and tasks. For accurate shelf-level estimation, we round each object's predicted height to the nearest shelf-level height, and a similar post-processing is conducted for the object orientation. 
% To add complexity, although we consider only four orientations: front, back, left and right, we discretize the possible orientations into $8$ possible options and round the predicted orientation to the nearest option. 
In our experiments, the object selection network was $100\%$ accurate, and the number of pixels wrongly classified was about $1\%$ of the complete image on average over $100$ random samples. The average error in predicting the height of the target placement after post-processing is around $8$~mm, and the mean error in the yaw angle of the predicted pose is $0.3$~rad.

% Average Pose Loss: 0.008775011206108865
% Average Orientation Loss: [0.01257151 0.         0.0210217  0.00975848]
% Average Object Loss: 0.0
% Average Segmentation Loss: 808.66

\subsection{Performance Evaluation: Diffusion with Guidance}

% \addressed{how many samples does the DM generate?}
The trained classifier-free conditional diffusion model and the score feasibility models are used to perform the reverse diffusion using the classifier-free guidance with and without feasibility score guidance. Experiments comparing the performance of both the methods are shown in~\autoref{fig:diffusionObjects} for a set of YCB Objects~\cite{calli2015ycb} and different scene-task scenarios where only $40$ candidate poses are sampled and top $10$ high-likelihood poses are selected. The comparison shows that while the classifier-free guidance is good enough to sample high-likelihood reorientation poses, the primary purpose of the feasibility score gradients is to reduce the variance in the pose generation and ensure high success probability. A numerical analysis of the overall success is shown and compared with the rejection sampling based baseline~\cite{wada2022reorientbot} in~\autoref{tab:performance}. 

\begin{table}[h]
\caption{Success evaluation of the proposed method as compared to the rejection sampling based baseline ReorientBot. The ReorientDiff algorithm was tested for more than 100 different scene task settings consisting of equal distribution of the selected objects and all the orientations. A task is considered a success if it is completed at-least once in 3 random seeds.}
\label{tab:performance}
\centering
\begin{tabular}{c|c|c|c}
\hline
\textbf{Method}               & \textbf{\begin{tabular}[c]{@{}c@{}}Success (\%)\\ Reorient\end{tabular}} & \textbf{\begin{tabular}[c]{@{}c@{}}Success (\%)\\ Place\end{tabular}} & \textbf{\begin{tabular}[c]{@{}c@{}}Success (\%)\\ Overall\end{tabular}} \\ \hline
Random   & 43.4  & 40.8  & 40   \\ 
ReorientBot   & 97.9  & 95.1  & 93.2   \\ 
\begin{tabular}[c]{@{}c@{}}ReorientDiff\\ (w/o Guide)\end{tabular}         & 97.4   & 92.3  & 90.8          \\ 
\textbf{ReorientDiff} & \textbf{98.9}   & \textbf{96.5}    & \textbf{95.2}  \\
\hline                                                          
\end{tabular}
\vspace{-1em}
\end{table}

The reorientation success percentage holds different relevance as compared to the baseline. The baseline does two step reverse rejection sampling where reorientation search is conducted over candidates which are feasible for placement, so there might be a scenario where there is no solution. For the case of ReorientDiff, the reorientation success measures the capability of the diffusion model to generalize to poses which ensure reorientability and scope for future placement. Higher reorientation success and lower placement success is an indication that the model is short-sighted and is giving importance to a single step success metric. From~\autoref{tab:performance}, we ensure high reorientability success along with better placement success. The overall success is based on the accurate placement of the object from the reoriented pose, and it represents the successful completion of a task. The metric is measured by calculating the difference between the desired and the pose after final placement.

% \begin{table}[h]
% \caption{(Incomplete) Ablation and Comparison Performance \chen{why is our planning time so long?}}
% \label{tab:performance}
% \centering
% \begin{tabular}{c|c|c|c}
% \textbf{Method} & \textbf{\begin{tabular}[c]{@{}c@{}}Success (\%)\\ Place\end{tabular}} & \textbf{\begin{tabular}[c]{@{}c@{}}Success (\%)\\ Overall\end{tabular}} & \textbf{\begin{tabular}[c]{@{}c@{}}Planning \\ Time (sec)\end{tabular}} \\ \hline
% ReorientBot           &         95.1  & 93.2 & \textbf{2.5}     \\ 
% ReorientDiff (w/o Guide) & 86.3 & 85.8 &  2.7 \\ 
% \textbf{ReorientDiff} & \textbf{95.1} & \textbf{93.7} & 5.3          
% \end{tabular}
% \end{table}

\begin{table}[h]
\caption{Success evaluation with different levels of discretization while sampling using ReorientDiff.}
% \chen{the performance is too sensitive to $K$. the result should be better if we use DEIS}}
\label{tab:kperformance}
\centering
\begin{tabular}{c|c|c|c}
\hline
\textbf{ReorientDiff K}               & \textbf{\begin{tabular}[c]{@{}c@{}}Success (\%)\\ Reorient\end{tabular}} & \textbf{\begin{tabular}[c]{@{}c@{}}Success (\%)\\ Place\end{tabular}} & \textbf{\begin{tabular}[c]{@{}c@{}}Success (\%)\\ Overall\end{tabular}} \\ \hline
$K = 10$   &  97.4  & 94.5  & 93.9   \\ 
\textbf{$K = 20$}   &  \textbf{98.9}  & \textbf{96.5}  & \textbf{95.2}   \\ 
% $K = 200$    & 97.2   & 93.2       & 92.9  \\ 
% $K = 50$ & 98.3  & 94.8 & 93.9          \\
% % $K = 100$ &     97.5         & 91.1  & 88.6 \\
\hline
\end{tabular}
\vspace{-1em}
\end{table}

\subsection{Performance Evaluation: K-Step Reverse Diffusion}

% \begin{table}[h]
% \caption{(Incomplete) Ablation and Comparison Performance \chen{why is our planning time so long?}}
% \label{tab:timeperf}
% \centering
% \begin{tabular}{c|c|c}
% \textbf{Method} & \textbf{\begin{tabular}[c]{@{}c@{}}Success (\%)\\ Place\end{tabular}} & \textbf{\begin{tabular}[c]{@{}c@{}}Planning \\ Time (sec)\end{tabular}} \\ \hline
% \textbf{ReorientDiff @ $K = 256$} & \textbf{95.1} & \textbf{5.3}  \\
% ReorientDiff @ $K = 200$ & TBD & 4.3                   \\
% ReorientDiff @ $K = 100$ & TBD & 2.5               \\
% ReorientDiff @ $K = 50$ & TBD & 1.5                
% \end{tabular}
% \end{table}

Sampling from a trained diffusion models is flexible and can be achieved using different levels of discretization between $x_K \sim \mathcal{N}(0, \mathbf{I})$ to meaningful reorientation poses. We perform the complete analysis for multiple values of the number of reverse denoising steps $K$ as shown in~\autoref{tab:kperformance}. ReorientDiff performs well with only 20 sampling steps.  

Following our analysis on performance, we explored the time consumption for the overall planning of a successful reorientation pose from a given scene and corresponding task information. We provide the recorded timings for all of our ablations and the baseline in~\autoref{tab:tperformance}.
\begin{table}[h]
\caption{Computational analysis of the planning time for ReorientDiff~($K=20$) with and without feasibility score guidance along with the baseline.}
\label{tab:tperformance}
\centering
\begin{tabular}{c|c}
\hline
\textbf{Method}               & \textbf{\begin{tabular}[c]{@{}c@{}}Planning \\ Time (sec)\end{tabular}} \\ \hline
ReorientBot   & 2.5 \\ 
ReorientDiff (w/o Guide)   & 0.3 \\ 
% ReorientDiff @ $K = 10$ & 5.3 \\
\textbf{ReorientDiff}         &      \textbf{1.05}\\ 
% % ReorientDiff @ $K = 200$    & 4.3 \\ 
% ReorientDiff @ $K = 50$ & 2.5\\
% ReorientDiff @ $K = 100$ & 5.3 \\
\hline
\end{tabular}
\vspace{-1em}
\end{table}

Our findings show that ReorientDiff leverages fast sampling strategies of FastDPM~\cite{kong2021fast} to recover from computationally heavy gradient calculations for reverse denoising steps. Without using the guidance from the feasibility-score models, classifier-free guidance requires even less time as compared to the baseline, ReorientBot, as shown in~\autoref{tab:tperformance}. Hence, from our visual and empirical analysis, ReorientDiff successfully proves that formulating the problem of reorientation as learning a conditional distribution is an efficient and scalable way to move towards more generalizable object manipulation.

\section{Conclusion}
\label{sec:conclusion}

Diffusion models are powerful generative models capable of modeling (conditional) distributions. Our proposed method ReorientDiff exploits the capabilities of such models to predict reorientation poses conditioned on a compact scene-task representation embedding containing information about the target object and its placement location. Further, the samples are refined using learned feasibility-score models to reduce uncertainty and ensure the success of the planned intermediate poses. With only $10$ candidate reorientation poses, we achieved an overall success rate of $95.2$\% across various objects. With the possible inclusion of point-cloud-based object representations~\cite{simeonov2023shelving}, such a method can generalize to a more diverse set of objects.
% We believe that with the incorporation of several high-order solvers~\cite{zhang2022fast} for solving the reverse diffusion sampling, we can drastically reduce the planning time without trading off the final performance. 
% We consider incorporating more efficient sampling schemes and better generalization performance for unseen objects and placement goals as a potential future work.

\bibliographystyle{ieeetr}
\bibliography{references}

\end{document}